\documentclass{article}
\usepackage{spconf,amsmath,graphicx,hyperref}

\title{
AEGIS-Net: Attention-guided Multi-Level Feature Aggregation for Indoor Place Recognition 
}
\name{Yuhang Ming$^{1,2}$*, Jian Ma$^{3}$*, Xingrui Yang$^{4}$, Weichen Dai$^{1,2}$, Yong Peng$^{1,2}$, Wanzeng Kong$^{1,2}$
\thanks{* Equal Contribution.}
\thanks{
The code is available at \url{https://github.com/YuhangMing/AEGIS-Net}.
}
\thanks{This research has been supported in part by the National Key Research and Development Program of China under Grant 2023YFE0114900.}}
\address{
$^{1}$School of Computer Science, Hangzhou Dianzi University, Hangzhou, China\\
$^{2}$Zhejiang Key Laboratory of Brain-Machine Collaborative Intelligence, Hangzhou, China\\
$^{3}$Unaffiliated\\
$^{4}$High-speed Aerodynamics Institute, CARDC
}
\begin{document}
%
\maketitle
\begin{abstract}
We present AEGIS-Net, a novel indoor place recognition model that takes in RGB point clouds and generates global place descriptors by aggregating lower-level color, geometry features and higher-level implicit semantic features. However, rather than simple feature concatenation, self-attention modules are employed to select the most important local features that best describe an indoor place.
Our AEGIS-Net is made of a semantic encoder, a semantic decoder and an attention-guided feature embedding. The model is trained in a 2-stage process with the first stage focusing on an auxiliary semantic segmentation task and the second one on the place recognition task.
We evaluate our AEGIS-Net on the ScanNetPR dataset and compare its performance with a pre-deep-learning feature-based method and five state-of-the-art deep-learning-based methods. Our AEGIS-Net achieves exceptional performance and outperforms all six methods.
\end{abstract}
%
\section{Introduction}
\label{sec:intro}
\vspace*{-8pt}
Place recognition allows autonomous robots to recognize a previously visited place in large environments. It is a popular research topic as it is crucial for global localization, preceding the 6 degree-of-freedom (DoF) pose estimation. When solving the place recognition task, it is commonly treated as a retrieval problem, which involves creating a global descriptor from local features and matching it with a database of known place descriptors.

\begin{figure}[t]
  \centering
  \includegraphics[width=0.45\textwidth]{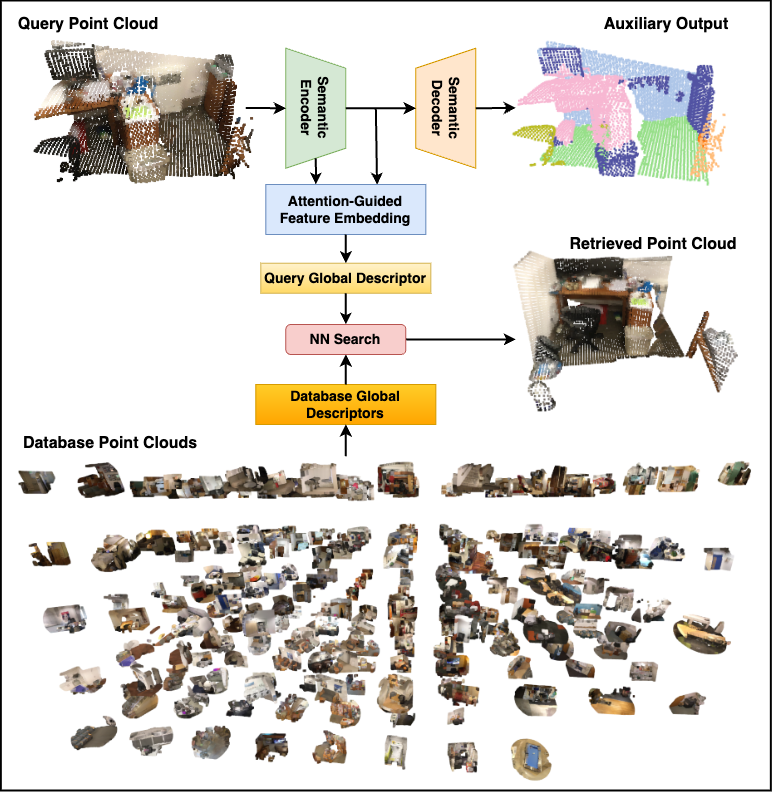}
  \vspace*{-2ex}
  \caption{Overview of the proposed AEGIS-Net. 
  }
  \label{fig::overview}
  \vspace*{-15pt}
\end{figure}

\begin{figure*}[t]
  \centering
  \includegraphics[width=0.98\textwidth]{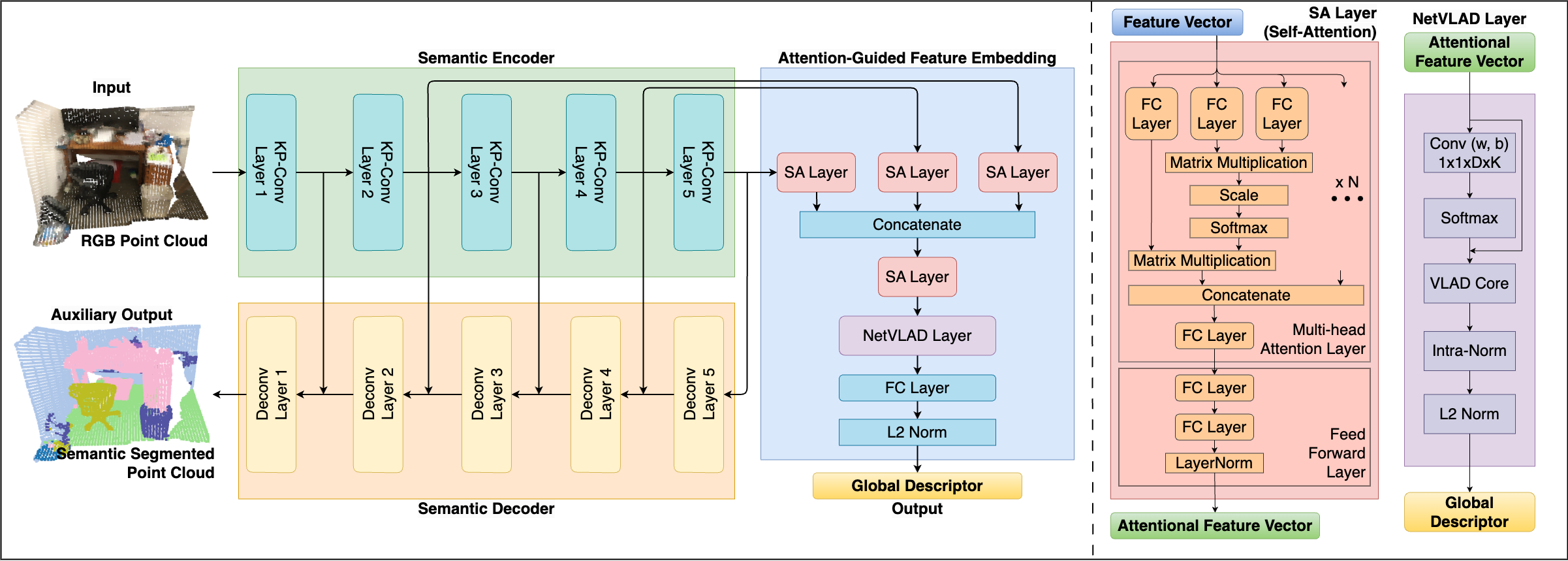}
  \vspace*{-2ex}
  \caption{The architecture of the proposed AEGIS-Net.}
  \label{fig::architecture}
  \vspace*{-3ex}
\end{figure*}

Place recognition in outdoor settings has garnered significant attention. 
Pioneered by NetVLAD~\cite{NetVLAD} with images input and PointNetVLAD~\cite{PointNetVLAD} with point clouds input, loads of end-to-end approaches have emerged with extraordinary performance like \cite{PatchNetVLAD, MinkLoc3D, garg2022ijrr}, along with 
substantial outdoor datasets, for example OxfordRobotCar~\cite{OxfordRobotCar} and CrossSeason~\cite{CrossSeason}.

However, indoor place recognition remains under-explored, presenting distinct challenges. For instance, sensors in indoor environments, like rooms, often capture limited scene portions, resulting in strong locality in the available information. Plus, indoor locations frequently exhibit similar structures and appearances, which challenges the discriminative power of 2-D data like RGB images or 3-D data like point cloud data alone. Combining these two, \cite{Sizikova2016ECCVW} employs a Siamese network for feature extraction from intensity and depth pairs while \cite{FDSLAM} enhances DH3D~\cite{DH3D} for indoor scenes by adding color as additional features to point cloud inputs. Exploiting the structural features, LCD~\cite{LCD} feeds both RGB-D frames and line clusters to the recognition network, ensuring the generated global descriptors retain structural information. On the other hand, SpoxelNet~\cite{SpoxelNet} focuses on the feature-manipulation by leveraging multi-level features of point clouds. It also tackles indoor occlusion issues by introducing the quad-view integrator.

In this paper, we focus on large-scale indoor place recognition and propose AEGIS-Net, short for AttEntion, color, Geometry and Implicit Semantics. 
Extended from our previous work CGiS-Net~\cite{cgisnet}, it is the first work to utilize self-attention to guide local feature selection, resulting in 
more discriminative global place descriptors. We propose that integrating semantic information with appearance and structural data can significantly enhance the performance of indoor recognition. For this, we've designed an approach, depicted in Fig. \ref{fig::overview}, that blends color, geometry and implicit semantic features with self-attention. Taking a cue from \cite{Schonberger2018CVPR}, we utilize an auxiliary semantic segmentation task to train the network that extracts semantics-enriched local features. The local features are then selected with self-attention (SA) layers and embedded into global descriptors.

We evaluate the proposed AEGIS-Net on the ScanNetPR dataset~\cite{cgisnet}, which is derived from the ScanNet dataset~\cite{ScanNet} and supports both point clouds and images inputs.
For comparison, a baseline model with traditional hand-crafted features \cite{SIFT, BoW}
and five prominent deep learning models~\cite{NetVLAD, PointNetVLAD, MinkLoc3D, FDSLAM, cgisnet}. The results 
demonstrates AEGIS-Net's superiority over renowned place recognition methods.
\vspace*{-10pt}

\section{Methodology}
\label{Sec::Method}
\vspace*{-12pt}
In this paper, we
follow our previous work~\cite{cgisnet} and use RGB point clouds as the network input to leverage color and geometry features. Regarding the semantics, a semantic encoder, which is trained separately on an auxiliary semantic segmentation task, is employed. Then, to balance the impact of color, geometry and implicit semantic features, SA layers are used to learn the weights adaptively before generating global descriptors with the NetVLAD layer.
\vspace*{-14pt}

\subsection{Network Architecture}
\vspace*{-8pt}
The architecture of the proposed AEGIS-Net is shown in Fig.~\ref{fig::architecture} (Left). It has three main parts: semantic encoder, semantic decoder and attention-guided feature embedding. Considering the varying point cloud densities in the indoor scenarios, we build the semantic encoder and semantic decoder on the advanced 3-D point cloud segmentation network KP-FCNN~\cite{KPConv}. However, the encoder-decoder architecture can be replaced with any point cloud segmentation networks. 
As for the attention-guided feature embedding, its core components are the SA layers and the NetVLAD layer.

The semantic encoder consists of 5 kernel point convolutional (KP-Conv) layers while the semantic decoder uses the same number of the nearest upsampling layers. In addtion, skip connections are used to pass information between the corresponding encoder-decoder layers. Since the KP-Conv mimics the behaviors of 2-D image convolutions, the earlier layers in the semantic encoder focuses on lower-level color and geometry features like corners, whereas later layers extracts more higher-level semantic features. For a deeper insight into behaviors of the KP-Conv layers, please check out the original KP-FCNN paper \cite{KPConv}.
To incorporate color, geometry and semantics features for place recognition, we use the features generated by the semantic encoder and the intermediate features extracted by the first 2 and 4 kernel point convolutional layers. Since the number of local features extracted by different layers varies a lot, these multi-level local features are first selected with SA layers respectively before concatenation. 
Then the combined local features undergo further enhancement with an additional SA layer. Finally, the enhanced representation is processed by a NetVLAD layer \cite{NetVLAD} to produce the global place descriptor. The detailed architectures of the SA layer and the NetVLAD layer are shown in Fig.~\ref{fig::architecture} (Right). For efficient retrieval, a subsequent fully-connected (FC) layer is added after the NetVLAD layer to reduce dimensions of the global descriptor.
\vspace*{-14pt}
\subsection{Self-Attention Layer}
\vspace*{-8pt}
The core of our proposed AEGIS-Net are the SA layers. As shown in Fig 2 (Right), each SA layer comprises a multi-head attention layer and a feed-forward layer. It allows the model to assign different weights to the local features and select the most important ones for the current place to generate the global descriptor.

Specifically, in each attention head, the attention is computed based on three vectors: Query ($\mathbf{Q}$), Key ($\mathbf{K}$) and Value ($\mathbf{V}$), which are obtained by applying three separate FC layers on the input feature matrix $\mathbf{X}$:
\begin{equation}
\mathbf{Q} = \mathbf{X} \mathbf{W}_Q; \; \mathbf{K} = \mathbf{X} \mathbf{W}_K; \; \mathbf{V} = \mathbf{X} \mathbf{W}_V 
\vspace*{-1ex}
\end{equation}
where $\mathbf{W}_Q$, $\mathbf{W}_K$, and $\mathbf{W}_V$ are the weight matrices of FC layers for the Query, Key and Value respectively. Then, the attention weights are computed using the softmax function:
\begin{equation}
\text{Attention}(\mathbf{Q}, \mathbf{K}, \mathbf{V}) = \text{softmax}\left(\frac{\mathbf{Q} \mathbf{K}^T}{\sqrt{d_K}}\right) \mathbf{V}
\vspace*{-1ex}
\end{equation}
where $d_k$ is the dimensionality of the Key vectors. The result is a weighted sum of the Value vectors where the weights represent the attention.
Then, the result computed from each attention head are concatenated together and passed through another FC layer for feature fusion. Finally, the output is stretched to the desired dimension with a feed forward layer.
\vspace*{-14pt}

\subsection{Multi-stage learning}
\vspace*{-8pt}

To force the encoder to learn actual scene semantics, we follow our previous work \cite{cgisnet} and adopt a 2-stage training process for our AEGIS-Net.
In the first stage, we train the encoder and decoder on the semantic segmentation task, thus, semantic encoder and semantic decoder. Unlike many other works which 
uses the explicitly segmented output for place recognition and localization \cite{Ramtoula2020ICRA, Ming2021IROS}
we favor the implicit features from the semantic encoder. Therefore, we call these features the implicit semantic features.

Then, the attention-guided feature embedding is trained with the encoder's weights fixed in the second stage.
Using the approach of PointNetVLAD~\cite{PointNetVLAD}, we employ metric learning with lazy quadruplet loss. The input of the model is a tuple of point clouds $\mathcal{T}=(P^{anc}, P^{pos}, P^{neg}, P^{*})$. Inside the tuple, $P^{anc}$ is the anchor point cloud representing the current place. $P^{pos}$ are the set of positive point clouds, which are taken from the same place as the anchor one but different viewpoints. $P^{neg}$ is the set of negative point clouds that are taken from different places than the anchor one, \textit{i.e.} from the same room but different place or from a complete different room. Finally, $P^{*}$ is a unique other negative point cloud that is negative to all previous point clouds in the input tuple. With the input tuple constructed, the lazy quadruplet loss can be computed as:


\vspace{-3ex}

\begin{equation}
\begin{split}
    \mathcal{L}_{LazyQuad}(\mathcal{T}) & = \underset{i,j}{\text{max}}([\alpha + \delta^{pos}_i-\delta^{neg}_j]_{+}) \\
    & + \underset{i,k}{\text{max}}([\beta + \delta^{pos}_i-\delta^{*}_k]_{+})
\end{split}
\vspace{-1ex}
\end{equation}
where $[\dots]_{+}$ represents the hinge loss with margins $\alpha$ and $\beta$. $\delta^{pos}_i$, $\delta^{neg}_j$, and $\delta^{*}_k$ are the Euclidean distances between respective point clouds.





\section{Experiments and Results}
\label{sec:exp}
\vspace*{-8pt}

\begin{figure*}[t]
  \centering
  \includegraphics[width=0.98\textwidth]{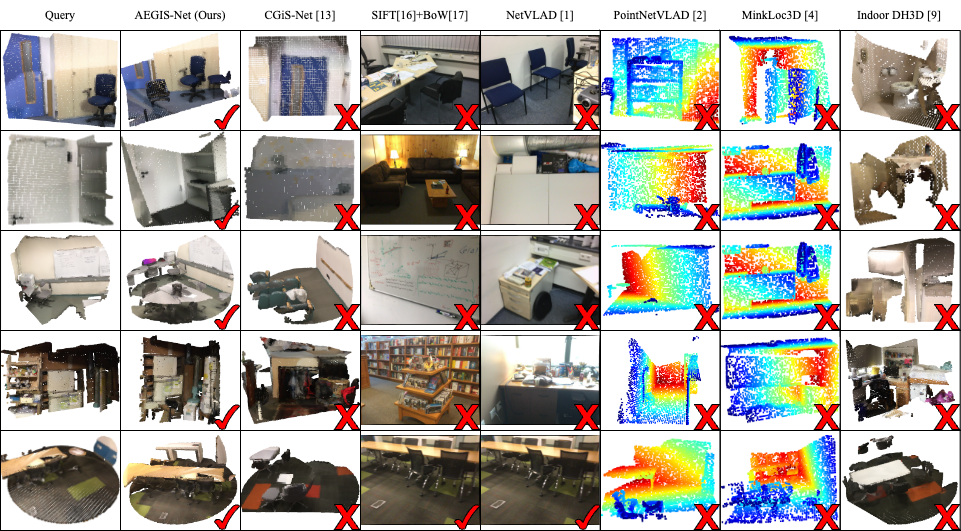}
  \vspace*{-2ex}
  \caption{Top-$1$ retrieval examples. Red checkmarks stand for succeeded retrievals and red crosses for failed ones. The query entries are visualized in RGB point clouds and the retrieved database entries are visualized in the same form as their inputs. }
  \label{fig::results}
  \vspace*{-3ex}
\end{figure*}

\textbf{Dataset:}
We use the ScanNetPR dataset~\cite{cgisnet}, which is derived from the famous ScanNet dataset~\cite{ScanNet}, to evaluate the proposed AEGIS-Net. ScanNetPR consists of 1,613 scans of 807 different rooms with the length varies according to the size of the room, the same as ScanNet. Overall, the dataset contains 35,102 training keyframes, which are selected from 1,201 scans of 565 rooms, 9,693 validation keyframes selected from 312 scans of 142 rooms and 3,608 testing keyframes, which covers the last 100 scans of 100 rooms. Each keyframe carries a RGB point cloud and a corresponding image to accommodate various input needs.

\textbf{Training Setup:}
In training stage 1, we follow the KP-FCNN~\cite{KPConv} paper and adopt stochastic gradient descent (SGD) optimizer to train the semantic encoder and semantic decoder for 50 epochs. All the hyperparameters in this training stage are exactly the same as those of the SLAM segmentation setup in KP-FCNN.


Then, in training stage 2, we first set the criterion to determine the positive and negative point clouds. Considering the size of typical rooms, two point clouds are positive if they share the same room ID and the distance between their centroids is less than $2m$, otherwise they are negative. However, we set the negative distance threshold to be $4m$ to maximize their differences. 
Additionally, constrained by the hardware memory limits, we opt for 2 positives and 6 negatives when constructing input tuples.
The attention-guided feature embedding is trained for 20 epochs using the Adam optimizer~\cite{adam} with an initial learning rate of 0.0001 and learning rate decay. Weight decay is applied to counter overfitting.  Mirroring typical SA layer and NetVLAD layer hyperparameters \cite{PointNetVLAD, MinkLoc3D, trans}, we choose 4 attention heads and 64 VLAD clusters. Trying to balance the representation power and the retrieval efficiency, we set the dimension of the final global descriptor to be 256. Finally, the margin parameters for the lazy quadruplet loss are set to $\alpha=0.5$ and $\beta=0.2$. 


\begin{table}[t]
\caption{Quantitative Results. Average Recall Rate (\%).}
\begin{center}
\begin{tabular}{|c||c|c|c|}
\hline
 Methods & R@1 & R@2 & R@3 \\
\hline
\hline
\textbf{AEGIS-Net} (Ours) & \textbf{65.09} & \textbf{74.26} & \textbf{79.06} \\
\hline
CGiS-Net \cite{cgisnet} & 61.12 & 70.23 & 75.06 \\ 
\hline
SIFT \cite{SIFT} + BoW \cite{BoW} & 16.16 & 21.17 & 24.38 \\
\hline
NetVLAD \cite{NetVLAD} & 21.77 & 33.81 & 41.49 \\
\hline
PointNetVLAD \cite{PointNetVLAD} & 5.31 & 7.50 & 9.99 \\
\hline
MinkLoc3D \cite{MinkLoc3D} & 3.32 & 5.81 & 8.27 \\
\hline
Indoor DH3D \cite{FDSLAM} & 16.10 & 21.92 & 25.30 \\
\hline
\hline
CGiS-Net-20 \cite{cgisnet} & 56.82 & 66.46 & 71.74 \\ 
\hline
AEGIS-Net (w/o attention) & 55.13 & 66.19 & 71.95 \\
\hline
\end{tabular}
\end{center}
\label{tab::compare}
\vspace{-2em}
\end{table}


\textbf{Results and discussions:}
Table~\ref{tab::compare} and Fig.~\ref{fig::results} showcase the average recall rates and top-1 retrievals of various place recognition methods repectively, with the first row in the table and second column in the figure highlighting the performance of our proposed AEGIS-Net. 
Remarkably, AEGIS-Net dominates across both quantitative and qualitative results
. Specifically, for quantitative result, our AEGIS-Net
records scores of \textit{65.09\%}, \textit{74.26\%}, and \textit{79.06\%} for top-1 recall (R@1), top-2 recall (R@2) and top-3 recall (R@3) respectively. When juxtaposed with CGiS-Net~\cite{cgisnet},
which is our previous work in the default setting,
AEGIS-Net displays notable superiority with a consistent \textit{4\%} improvement, suggesting that the enhancements introduced in AEGIS-Net have significantly bolstered its performance. 
Apart from that, an image-based traditional method which combines SIFT~\cite{SIFT} with bag-of-words (BoW)~\cite{BoW} is chosen as a baseline model. Following that, more advanced learning-based methods whose official implementations are available are chosen for comparison. In particular, NetVLAD~\cite{NetVLAD} with images input, PointNetVLAD~\cite{PointNetVLAD} and MinkLoc3D~\cite{MinkLoc3D} with point clouds input, and indoor DH3D~\cite{FDSLAM} with RGB point clouds input are re-trained and evaluated on the ScanNetPR dataset with their default parameters. 

Notably,
while NetVLAD~\cite{NetVLAD} exhibits a marked improvement over the baseline, indicating the progression of early deep learning models, it still lags considerably behind AEGIS-Net. 
On the other hand, despite PointNetVLAD~\cite{PointNetVLAD}, MinkLoc3D~\cite{MinkLoc3D} and DH3D~\cite{DH3D} prevailed in the outdoor settings,
they surprisingly underperform in indoors, even with indoor modification~\cite{FDSLAM}. 
This vast disparity in performance demonstrates that using color or geometry features along are not enough for indoor place recognition, further accentuating the efficacy and robustness of our AEGIS-Net, setting it apart as a leader in the realm of indoor place recognition.


\textbf{Ablation study:} To further prove the effectiveness of the attention mechanism of our AEGIS-Net, ablation experiments are performed with the results shown at the bottom two rows of Table~\ref{tab::compare}. From the efficiency perspective, our previous work, CGiS-Net converges after 60 epochs of training, and the training process takes around 3 weeks. When limiting the training epochs to 20, CGiS-Net still needs 7 days at the cost of drastic drop in the performance, shown in row ``CGiS-Net-20''.
While our AEGIS-Net reaches convergence by the 20th training epoch, and the training time is largely reduced to 4 days.
Beyond training speed, the use of attention in the model also delivers superior performance. 
Precisely, the attention-equipped model surpasses its attention-absented counterpart by \textit{10\%} for top-1 recall and \textit{8\%} for top-2 and top-3 recall, as shown in the row ``AEGIS-Net (w/o attention)''.
This improvement isn't just a testament to the effectiveness of the attention mechanism but also indicates its role in assisting the network to select more discriminative and relevant features for place description.
\vspace*{-3pt}

\section{Conclusion}
\label{sec::con}
\vspace{-2pt}
We have presented AEGIS-Net for indoor place recognition which 
is capable of selecting color, geometry and semantics features that best describe a particular place in the indoor scenes with attention-guidance.
The network is trained and evaluated on the ScanNetPR dataset with superior performance compared to state-of-the-art learning-based methods.

\end{document}